 \documentclass[pmlr,twocolumn,10pt]{jmlr} % W&CP article

\usepackage{booktabs}
\usepackage{siunitx}

\usepackage{boldline}
\usepackage{caption}

\theorembodyfont{\upshape}
\theoremheaderfont{\scshape}
\theorempostheader{:}
\theoremsep{\newline}

\jmlrvolume{}
\jmlryear{2023}
\jmlrsubmitted{}
\jmlrpublished{}
\jmlrworkshop{Machine Learning for Health (ML4H) 2023} % W&CP title

\title[RL for measurement scheduling]{Measurement Scheduling for ICU Patients with Offline Reinforcement Learning}

\author{%
\Name{Zongliang Ji} \Email{jerryji@cs.toronto.com}\\
\addr University of Toronto, Canada
\AND
\Name{Anna Goldenberg} \Email{anna.goldenberg@utoronto.ca }\\
\addr University of Toronto, Canada
\AND
\Name{Rahul G. Krishnan} \Email{rahulgk@cs.toronto.edu }\\
\addr University of Toronto, Canada
}

\usepackage{soul}
\usepackage{amsmath}
\usepackage[ruled]{algorithm2e} % For algorithms
\newcommand{\nosemic}{\SetEndCharOfAlgoLine{\relax}}% Drop semi-colon ;
\begin{document}

\maketitle
\vspace*{-3\baselineskip}
\begin{abstract}
Scheduling laboratory tests for ICU patients presents a significant challenge. Studies show that 20-40\% of lab tests ordered in the ICU are redundant and could be eliminated without compromising patient safety. Prior work has leveraged offline reinforcement learning (Offline-RL) to find optimal policies for ordering lab tests based on patient information. However, new ICU patient datasets have since been released, and various advancements have been made in Offline-RL methods. In this study, we first introduce a preprocessing pipeline for the newly-released MIMIC-IV dataset geared toward time-series tasks. We then explore the efficacy of state-of-the-art Offline-RL methods in identifying better policies for ICU patient lab test scheduling. Besides assessing methodological performance, we also discuss the overall suitability and practicality of using Offline-RL frameworks for scheduling laboratory tests in ICU settings.

\end{abstract}

\begin{keywords}
Offline Reinforcement Learning,  Patient Measurement Scheduling, MIMIC-IV
\end{keywords}

\section{Introduction}
Laboratory tests are a pivotal component in guiding doctors' decisions regarding patient care. Based on these test results, clinicians diagnose conditions, determine appropriate treatments, and monitor the progress of patients' clinical states. The ordering of laboratory tests is an integral aspect of daily medical practice. 

However, laboratory tests can be expensive. Most of them necessitate blood draws or other invasive procedures, which have implications for the patients. Frequent testing can result in significant blood loss, sleep disruptions, and general discomfort. Current medical practices often entail the ordering of redundant lab tests. This not only elevates healthcare costs but also places a strain on both patient well-being and hospital resources (\citeauthor{feldman2009managing,badrick2013evidence}). Moreover, research indicates that routine blood tests don't necessarily enhance diagnostic accuracy (\citeauthor{iosfina2013implementation, pageler2013embedding}). In fact, ordering unnecessary invasive laboratory tests could exacerbate a patient's condition (\citeauthor{berenholtz2004eliminating, salisbury2011diagnostic}).

Due to the intricate challenges of caring for ICU patients, it becomes increasingly challenging for clinical teams to discern truly essential lab tests. Patients in the ICU, typically being in critical conditions, necessitate more laboratory tests than those in other wards. For instance, at Sunnybrook Hospital, while ICU patients constitute only 5\% of the hospital population, they account for 26\% of daily laboratory tests. Furthermore, it has been observed that between 20-40\% of these tests could be eliminated without jeopardizing patient safety.
Given that the majority of patient information during their ICU stay is digitally recorded, there's an opportunity to develop a computer-aided agent. This agent would propose daily laboratory test orders based on the patient's current status and medical history. By implementing an agent that suggests optimal test ordering policies proactively, we could enhance patient sleep quality, mitigate the risks of anemia, and minimize discomfort from venipuncture. Beyond patient comfort, the financial and environmental benefits are significant. Laboratory tests cost approximately \$1 billion annually in Ontario, and the use of lab specimen tubes contributes substantially to the carbon footprint. Additionally, such a computer agent can alleviate the cognitive load on clinicians and mitigate pressures associated with clinical ranking and responsibilities.

Reinforcement Learning (RL) is a compelling approach for training agents to observe the current state of an environment and take actions that maximize future rewards. This makes RL particularly apt for applications like creating an agent to guide ICU patient lab test scheduling. Many researchers have harnessed RL to develop policies for patient treatment plans (\citeauthor{ShamimNemati2016OptimalMD, AniruddhRaghu2017ContinuousSM, NiranjaniPrasad2017ARL, JosephFutoma2018LearningTT}). Specifically, two studies (\citeauthor{cheng2018optimal, chang2019dynamic}) have addressed the ICU patient measurement scheduling challenge within an offline-RL setting.

Since the study by \citeauthor{chang2019dynamic}, there has been a surge in offline-RL literature and the availability of ICU patient datasets. 
In this work, our aim is to assess the performance of state-of-the-art Offline-RL techniques in establishing efficient policies for ICU patient lab test scheduling. 
Our primary contributions encompass: (1) reproducing \citeauthor{chang2019dynamic}'s work in PyTorch (\citeauthor{paszke2019pytorch}) and adapting it to the MIMIC-IV (\citeauthor{johnson2023mimic}) dataset with preprocessing aligned to \citeauthor{Harutyunyan2019} for time-series tasks; (2) investigating a broad array of modern RL algorithms such as Behavior Cloning, Dueling-DQN (\citeauthor{jiang2016doubly}), CQL (\citeauthor{kumar2020conservative}), and IQL (\citeauthor{kostrikov2021offline}) for patient measurement scheduling across both MIMIC datasets; and (3) reevaluating the framing by \citeauthor{chang2019dynamic} and discussing the clinical implications of RL-based lab test scheduling.

% Our contributions can be described in three folds: first, we reproduced \citeauthor{chang2019dynamic}'s work in PyTorch (\citeauthor{paszke2019pytorch}) and applied to MIMIC-IV (\citeauthor{johnson2023mimic}) dataset. Then, we present our studying on a diverse set of modern RL algorithms for the same framing of measurement scheduling problem. Lastly, we revisit whether \citeauthor{chang2019dynamic}'s framing is the right way forward.
% We delve into the practicality of applying RL to patient measurement scheduling by testing various offline-RL methods on both prior and newly released datasets. Initially, we outline a preprocessing pipeline for the MIMIC-IV \citeauthor{johnson2023mimic} dataset, structured for time-series tasks in line with \citeauthor{Harutyunyan2019}. Subsequently, we reproduce \citeauthor{chang2019dynamic}'s work in PyTorch (\citeauthor{paszke2019pytorch}), and present the performance metrics of Behavior Cloning, Dueling-DQN (\citeauthor{jiang2016doubly}), CQL (\citeauthor{kumar2020conservative}), and IQL (\citeauthor{kostrikov2021offline}) on both the MIMIC-III and IV datasets (\citeauthor{johnson2016mimic, johnson2023mimic}) Our findings then inform a discussion on the clinical practicalities of scheduling lab tests.

%\footnote{Our code will be released on Github after review process.}. 

\vspace*{-1.5\baselineskip}
\section{Methods}

In our study, we cast the measurement scheduling problem within the context of the offline off-policy deep q-learning setting in RL, following the approach of \citeauthor{chang2019dynamic}. Beyond the dueling-DQN (\citeauthor{jiang2016doubly}), we also evaluate the performance of Behavior Cloning, CQL (\citeauthor{kumar2020conservative}), and IQL (\citeauthor{kostrikov2021offline}).  
\vspace*{-1\baselineskip}
\subsection{Dataset}
The Medical Information Mart for Intensive Care (MIMIC, \citeauthor{johnson2016mimic, johnson2023mimic}) is a publicly available ICU patient database, resulting from a collaboration between MIT and Beth Israel Deaconess Medical Center. This comprehensive dataset serves as a cornerstone for healthcare research, containing a myriad of data points from demographics to medications, and is instrumental for predictive modeling and policy examination.

MIMIC-III (\citeauthor{johnson2016mimic}) covers ICU patient data from 2001-2012, with the subsequent MIMIC-IV (\citeauthor{johnson2023mimic}) extending this range to 2019, providing a richer resource for contemporary research.
Adhering to the widely accepted MIMIC-III preprocessing methodology (\citeauthor{Harutyunyan2019}), we introduce a corresponding pipeline for MIMIC-IV. Alongside cleaning, extracting, and preprocessing the ICU patient time-series data, our approach also identifies the comorbidity associated with patient diagnoses. Detailed insights into our pipeline are elaborated upon in Appendix \ref{apd:first}.

\begin{algorithm2e}
\caption{Training Dueling DQN}
\label{alg:pi}
\SetKwProg{generate}{Function \emph{generate}}{}{end}
\SetKwRepeat{Do}{do}{while}
%\SetKwProg{generate}{Function \emph{generate}}{}{end}
\KwIn{Pretrained LSTM model $f$, patient ICU stay database $D = \{X^1, X^2, ..., X^N\}$, patient ICU stay's trajectory length $T^i$} 
%Lab tests ordered $A' \subseteq A$ and their values $X_{inv}^t = [x_1^t, ..., x_m^t ] $ between $t$ to $t+1$, }
\KwOut{DQN model $Q_\theta$}
\nosemic $R \gets  \emptyset$ \text{// Initialize experience replay buffer $R$}\;
\For{$X^i \text{ in } D$ }{
\For{$t=1$ \KwTo $T^i$}{
$E^i \gets$ get experience for ICU stay $X^i$ at $t$  \;
Store $E^i$ in $R$
}}
\While{$L$ is not converged}{
$E \sim R$ \;
$(s, a, r, s') \gets E$ \;
$Q_{target}(s,a,s') = r(s,a) + \gamma \mathop{max}_{a' \notin s'} Q_\theta(s',a')$\;
minimize $L = [Q_\theta(s,a) - Q_{target}(s,a,s')]^2$\;
Update policy of $E$ in $R$ using $L$
}
\end{algorithm2e}
\vspace*{-1.5\baselineskip}

\subsection{MDP Formulation}
To address this challenge with reinforcement learning, we begin by framing the task of discerning the optimal lab test ordering policy as a Markov Decision Process (MDP). For each ICU stay or patient admission, we define:
(1) A state space $\mathcal{S}$ where the patient's physiological state at time $t$ is represented as $s_t \in \mathcal{S}$.
(2) An action space $\mathcal{A}$ encompassing all potential lab tests a clinician may opt for at time $t$, denoted as $a_t \in \mathcal{A}$.
(3) An undetermined transition function $\mathcal{P}_{sa}$ that signifies the shift in the patient's health status.
(4) A reward function $r_t$ which mirrors the observed clinical feedback from the action $a_t$ executed at time $t$.

The primary goal of the RL agent is to derive an optimal policy $\pi^* : \mathcal{S} \rightarrow \mathcal{A}$ that elevates the expected discounted cumulative reward throughout an admission:
$$\pi^* = \underset{\pi}{\mathrm{argmax}} \mathop{{}\mathbb{E}} \left[ \sum_{t=0}^T \gamma^t r_t(s_t, a_t) | \pi  \right], $$
where $T$ designates the admission duration, and $\gamma$ is the discount factor.

Given that the transition function is unknown, and interacting with and gathering data from the environment (i.e., ICU patients) is infeasible, our problem is classified within the offline and off-policy RL framework. Off-policy RL methods typically employ a state-action value function, or Q-function, denoted as \( Q(s, a) \). This represents the discounted returns obtained by starting from state \( s \) and action \( a \), and subsequently adhering to policy \( \pi \). Offline RL operates using pre-existing data without necessitating further data collection. Consistent with many offline RL approaches, our methodology focuses on minimizing the temporal difference error, as defined by the subsequent loss:
\begin{equation} \label{ref:eq1}
    L(\theta) = \mathop{{}\mathbb{E}}_{(s,a,s') \sim \mathcal{D}} [( r(s,a) +  \mathrm{max}_{a'} Q_{\hat{\theta}}(s,a) - Q_{\theta}(s,a))^2]
\end{equation}
Here, \( \mathcal{D} \) denotes the dataset, \( Q_{\theta}(s,a) \) is a parameterized Q-function, \( Q_{\hat{\theta}}(s,a) \) is a target network, and the policy is expressed as \( \pi(s) = \mathrm{argmax}_a Q_\theta(s,a) \).

Dataset \( \mathcal{D} \) encompasses tuples \( (s,a,r,s') \) labeled as RL experiences. Within our measurement scheduling paradigm, each tuple encapsulates the state, action, and reward at time \( t \), as well as the subsequent state at time \( t+1 \) of a patient's ICU tenure. We utilized \citeauthor{chang2019dynamic}'s approach, first modeling patient trajectories via an LSTM (\citeauthor{hochreiter1997long}). Then, we crafted the patient state by melding the LSTM model's hidden representation (\( h \in \mathbb{R}^{256} \)) with the patient's historical measurement (\( m \in \{0,1\}^{39} \)) for each time step, yielding a state vector of dimension 295. The action \( a_t \in \{0, 1\}^{39} \) manifests as a binary vector, corresponding to the 39 feasible measurements under consideration.

The reward \( r \) is derived from the reward function:
$$r(s_t, a_t) = \Delta p - \lambda * c(a_t) .$$ 
Here, \( \Delta p \) signifies the probability difference (or information gain) of the patient trajectory model between times \( t \) and \( t+1 \). Meanwhile, \( c(a_t) \) represents the action cost, defined by the number of ordered measurements at time \( t \). Comprehensive details on generating RL experiences from ICU patient stays can be located in Appendix \ref{apd:second}.

\subsection{Policy Training and Evaluation}

For all the methodologies under consideration, we minimize the temporal difference error, as delineated in equation \ref{ref:eq1}. We employ the dueling-DQN network (\citeauthor{jiang2016doubly}) as our \( \theta \).

\textbf{Training:} Leveraging our preprocessed RL experience dataset in conjunction with the dueling-DQN setup, our objective is to determine the optimal policy \( Q \). The training mechanism of the Dueling DQN is elucidated in Algorithm \ref{alg:pi}. When implementing Behavior Cloning (BC), rather than computing the MSE loss between \( Q_\theta \) and \( Q_{target} \), we compute the Cross Entropy loss between the output of the Q-function and the action. This transformation reframes the problem as a multi-class classification challenge. In the context of CQL (\citeauthor{kumar2020conservative}), we introduce a soft target model update, while also integrating the CQL loss. For IQL (\citeauthor{kostrikov2021offline}), we adapt our training regimen by optimizing the upper expectile value function and subsequently utilize the output of this value function as the \( Q_{target} \), in alignment with the stipulated method.

\textbf{Off-policy Policy Evaluation:} Upon refining our trained policy \( Q_{\theta} \), we aim to gauge its efficacy relative to our stated objectives. Nonetheless, we encounter the predicament of the absence of clinicians to critique the lab tests ordering decisions that have been made. Moreover, there's no mechanism to witness counterfactual scenarios (e.g., understanding a patient's condition when specific tests were not prescribed) which are pivotal for evaluating our policy.

In response to this challenge, we harness the Off-Policy Policy Evaluation (OPPE) as the exclusive metric to appraise our trained policy against historical data. Our strategy is to cultivate a regression-based model \( \phi \) to approximate the values of both the physician's policy and our inculcated policies by leveraging data acquired by physicians.
Explicitly, we train \( \phi: s \times a \rightarrow \Delta p \), a model that associates the state-action pair to the metric of information gain, tantamount to the probability shifts of model \( f \). At each instance \( t \), we merge the latent state \( h_t \) of model \( f \) with a multi-hot representation of the actions \( a_t \) executed at time \( t \), and task \( \phi \) to generate the probability variations \( \Delta p \) for model \( f \) between instances \( t \) and \( t+1 \). By determining the information gain across all ICU stays, we can deduce the aggregate information gain \( G \) of a specified policy. A more granular exposition of OPPE is available in Appendix \ref{apd:third}.

% \vspace*{-1.\baselineskip}

% \begin{table}[htbp] %\label{tab1}
% %\centering
% \floatconts
%   {tab:igncost}
%     {\caption{Comparison of Different Offline-RL Methods, $G$ is the average cumulative information gain of policies and $C$ is the average policy cost.}
%     \vspace*{-1.5\baselineskip}}
%     {
%     \resizebox{\columnwidth}{!}{
    
%     \begin{tabular}{|c|ll|ll|}
%         \hline
%         \textbf{Method} & \textbf{G (III)} & \textbf{C (III)} & \textbf{G (IV)} & \textbf{C (IV)}\\
%         \hline
%         Physician & 1                     & 1          & 1   & 1        \\
%         \hline
%         Random & 1                       & 1         & 1  &   1       \\
%         BC    & 1                        & 1          & 1 & 1\\
%         \hline
%         DDQN  & 1                        & 1          &1  &  1       \\
%         CQL   & 1                        & 1          &1  &   1       \\
%         IQL   & 1                        & 1          &1 &    1    \\
%         \hline
%     \end{tabular}}
%     }
%     \vspace*{-1.5\baselineskip}
% \end{table}
\vspace*{-0.8\baselineskip}

\section{Experiment Results}

% We present a comparative evaluation of cumulative information gain and relative costs across Behavior Cloning (BC), dueling-DQN (DDQN \citeauthor{jiang2016doubly}), CQL (\citeauthor{kumar2020conservative}), and IQL (\citeauthor{kostrikov2021offline}). For consistency, the same DDQN network structure was employed across all methods, differing only in their training approaches.

% Guided by the experimental paradigm of \citeauthor{chang2019dynamic}, each method was exposed to a spectrum of learning rates, cost coefficients, and random seeds to uncover the optimal policy for measurement scheduling. Post-training, every derived policy was assessed on a separate test set to compute its cumulative information gain ($G$) and policy cost ($C$), spanning both MIMIC-III and MIMIC-IV datasets (\cite{johnson2016mimic, johnson2023mimic}).
We present a comparative evaluation of cumulative information gain and relative costs across Behavior Cloning (BC), dueling-DQN (DDQN \cite{jiang2016doubly}), CQL (\cite{kumar2020conservative}), and IQL (\cite{kostrikov2021offline}). For consistency, the same DDQN network structure was used across all methods, with differences only in their training approaches.
Following the experimental paradigm of \citeauthor{chang2019dynamic}, each method was exposed to a range of learning rates, cost coefficients, and random seeds to identify the optimal policy for measurement scheduling. After training, each derived policy was evaluated on a separate test set to compute its cumulative information gain ($G$) and policy cost ($C$). These evaluations spanned both the MIMIC-III and MIMIC-IV datasets (\cite{johnson2016mimic, johnson2023mimic}).
\vspace*{-1.1\baselineskip}
\begin{figure}[h]
\floatconts
  {fig:m3}
  {
  \vspace*{-1.5\baselineskip}
  \caption{Evaluation of various policies in MIMIC3}}
  {%
    \includegraphics[width=1\linewidth]{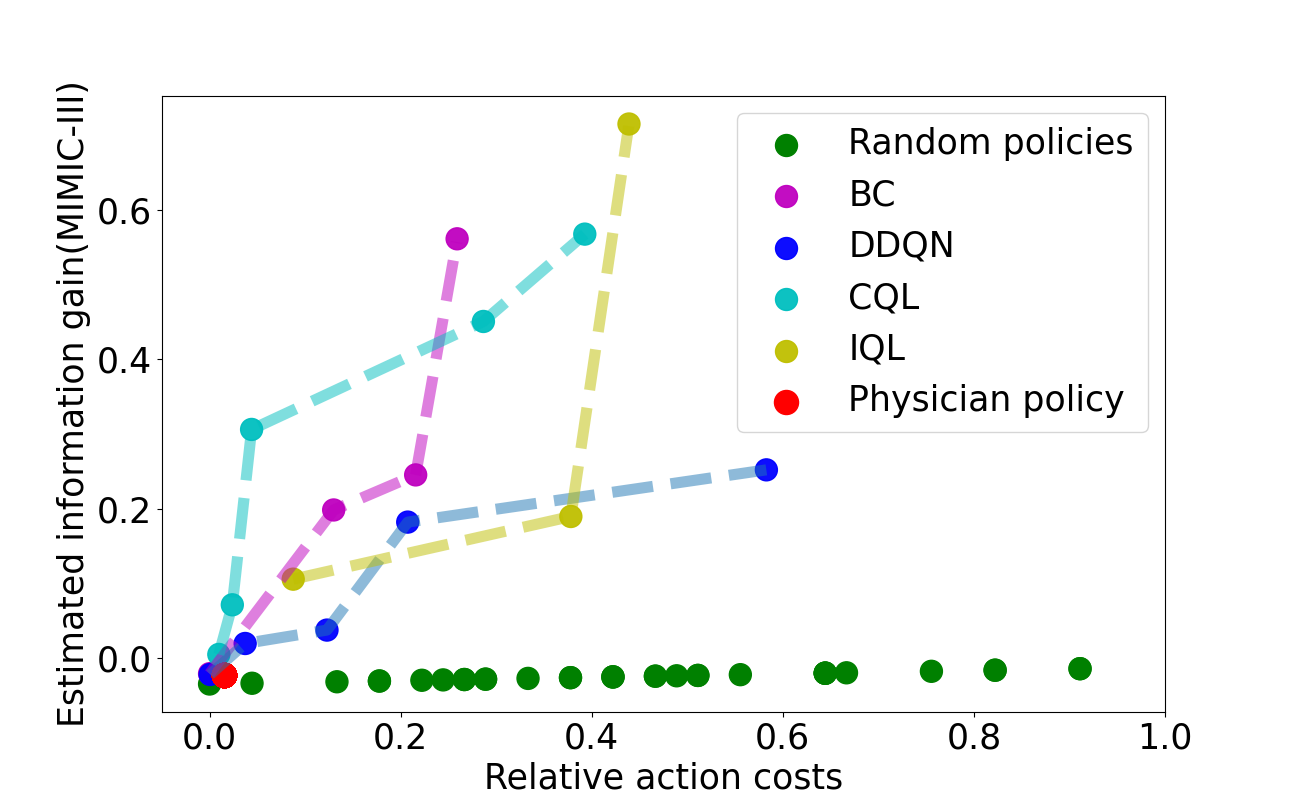}
  }
\end{figure}

\vspace*{-1.4\baselineskip}
\begin{figure}[htbp]
\floatconts
  {fig:m4}
  {\vspace*{-2\baselineskip}
  \caption{Evaluation of various policies in MIMIC4}}
  {\includegraphics[width=1\linewidth]{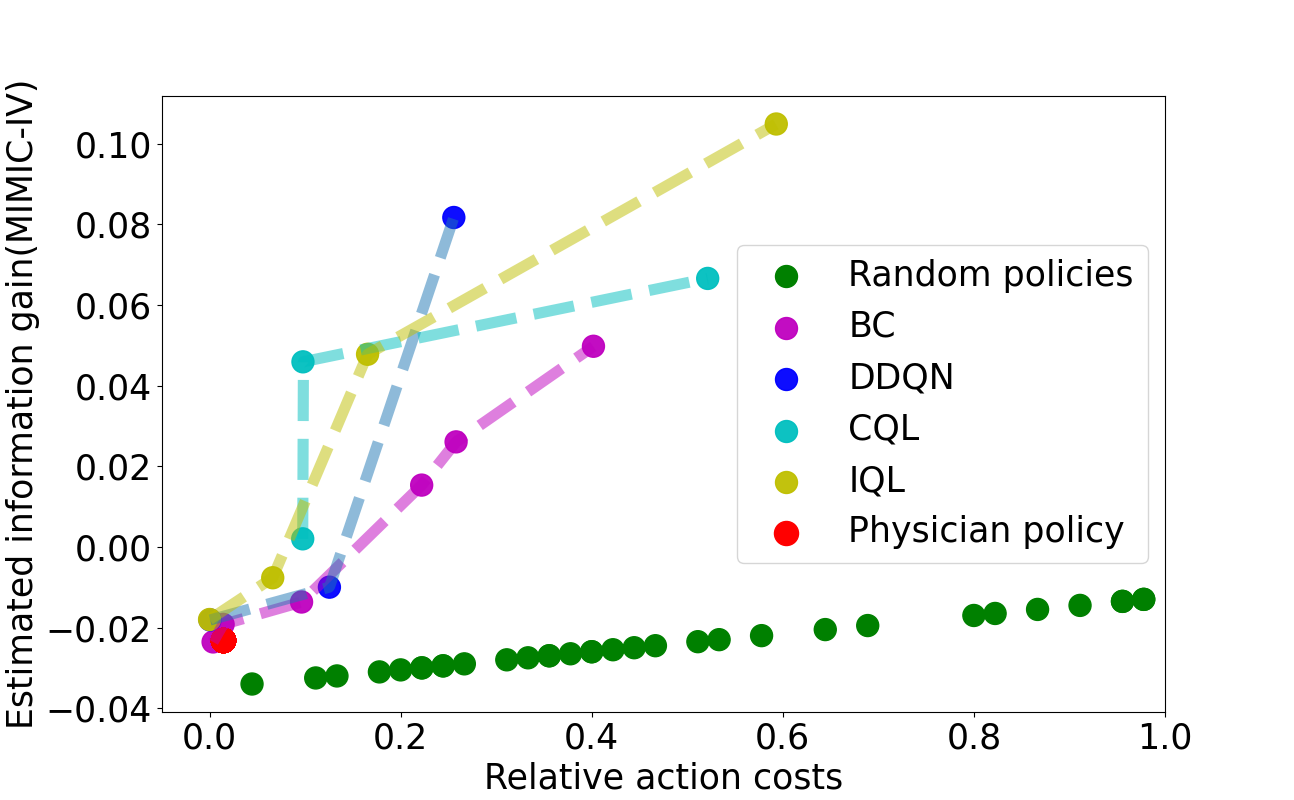}
  }
\end{figure}

The pinnacle policies from each method are depicted in Figure \ref{fig:m3} and \ref{fig:m4}. The desirability of a policy is gauged by its ability to offer a higher information gain at a reduced cost.

%\vspace*{-0.8\baselineskip}
%\vspace*{-0.3\baselineskip}

From our analysis, several key insights emerged. When we included the physician's policy as a reference, every method demonstrated superiority over random choices. Every approach displayed the capability to discover policies that boost information gain while minimizing costs, surpassing the physician's benchmarks. Notably, Behavior Cloning (BC) mirrored the performance metrics of IQL and CQL across both databases and outdid DDQN in the MIMIC-III dataset in terms of information gain. Since the two datasets (III \& IV) differed in patient cohorts and the missingness of measurement values during patient ICU stays, we could not ascertain why one method outperformed in one dataset but not in the other. CQL exhibited the best performance across both datasets. However, our results did not decisively indicate the superiority of one method over the others.
A more comprehensive discussion on average cumulative information gain, policy cost distributions, and additional analyses can be found in Appendix \ref{apd:fourth}.

\vspace*{-0.8\baselineskip}
\section{Discussion and Future Work}

Though a plethora of new offline-RL literature has been published since \citeauthor{chang2019dynamic} proposed applying offline and off-policy deep-RL to find the optimal policy for measurement scheduling, no studies have compared time-series based policies with RL-trained policies for this specific problem.
While all the RL methods operate under the assumption of a Markov Decision Process (MDP), which requires the memorylessness property, the current status of an ICU patient might not be entirely encapsulated by the present state. This state might be influenced by past treatments, the patient's status, or measurements from earlier time points.
From our results using the Behavior Cloning method, essentially a multi-hot classification, we deduced that RL methods aren't notably superior to BC for measurement scheduling. Currently, policies are evaluated based on the information gain (probability difference) of the patient trajectory model. However, this metric of information gain is somewhat nebulous for clinicians in their daily operations. If a model could predict exact values with confidence based on a patient's history, clinicians could make decisions on test ordering based on anomalies detected in the model output. Consequently, a future direction is to explore metrics with deeper clinical relevance when evaluating trained policies.

Presuming we attain a trained agent for ICU measurement scheduling, its deployment could pose challenges. To curate a meaningful offline RL dataset, the current agent operates on a half-hour interval between states. But in a non-emergency setting, clinicians typically order lab tests daily. Beyond this time discrepancy, some vital signals are incorporated into the action set. In the ICU, vital readings and other non-invasive measurements can be continuously captured without endangering patients and incurring extra costs, yet these vital signals dominate the RL dataset. Another avenue for future exploration is the creation of policies better tailored to real-world clinical scenarios.

\newpage

% \acks{This research is supported by CIFAR.}

\bibliography{reference}

\newpage

\appendix

\section{MIMIC Dataset: Preprocessing and Feature Selection}\label{apd:first}

MIMIC (Medical Information Mart for Intensive Care) is a publicly available database of de-identified electronic health records (EHRs) from patients admitted to the Beth Israel Deaconess Medical Center (BIDMC) in Boston, Massachusetts. MIMIC-IV is one of the largest and most comprehensive critical care databases available, containing data from over 300,000 hospital admissions between 2008 and 2019. It is released in 2021 and updated to a complete version in 2022.  

The MIMIC-IV dataset includes a wide range of clinical data, such as vital signs, laboratory test results, medication orders, procedures, diagnoses, and demographic information. The data is collected from various sources, including bedside monitors, electronic medical records, and nursing notes, among others. The data is stored in a relational database format, with each record corresponding to a specific patient encounter.   

One of the unique features of the MIMIC-IV dataset is the inclusion of waveform data, which provides high-resolution time-series data for various physiological signals, such as electrocardiograms (ECGs), arterial blood pressure, and respiratory signals. This waveform data can be used for advanced signal processing and machine learning applications, such as predicting patient outcomes and detecting abnormalities. The MIMIC-IV dataset also includes extensive clinical phenotyping, which involves identifying specific patient subgroups based on clinical characteristics and outcomes. This phenotyping can be used to develop and test predictive models for various clinical outcomes, such as mortality, length of stay, and readmission rates. 

To ensure patient privacy and confidentiality, the MIMIC-IV dataset is de-identified and follows the Health Insurance Portability and Accountability Act (HIPPA). It is released under a data use agreement, which requires users to follow strict guidelines for data security and ethical use. However, access to the dataset is free for researchers and clinicians who agree to these terms. 

Overall, the MIMIC-IV \citeauthor{johnson2020mimic} is a valuable resource for developing and testing predictive models, evaluating interventions, and improving ICU patient outcomes. Its previous version of the MIMIC-III dataset is well-studied and well-known for intensive care machine learning problems. With the success of its predecessor, the MIMIC-IV (\citeauthor{johnson2020mimic}) was just released and has not been fully explored.

\subsection{Preprocessing Pipeline for MIMIC-IV dataset}

We develop a set of Python scripts that preprocess and aggregate the MIMIC-IV (\citeauthor{johnson2020mimic}) raw data from relational database format into a format that can be utilized by deep-learning community.  
For developing our preprocessing procedure, we followed and extended a prior work bench-marking the MIMIC-III (\citeauthor{johnson2016mimic}) with Python (\citeauthor{Harutyunyan2019}). To our knowledge, we are the first to conduct this time-series preprocessing for MIMIC-IV (\citeauthor{johnson2020mimic}) dataset.

We first create a folder indexed by patient subject identification number and extract each patient's raw admission, ICU stays, diagnoses, and laboratory, input/output events information and saved into each patient folder. We then validate the extracted value and unify the missing values obtained from the raw data for each patient. After this step, we prepare the patient ICU stay into time-series data with episodes by event time stamps and store each episode's outcome (mortality, length of stay, diagnoses) in a separate file.  
To reproduce the work done by \citeauthor{chang2019dynamic}, we also generate a script to convert each patient's diagnoses codes into a multi-hot time-invariant features.  

We select 38 static demographic and clinical features: Age, Gender, Ethnicity, congestive heart failure, cardiac arrhythmias, valvular disease, pulmonary circulation, peripheral vascular, hypertension, paralysis, other neurological, chronic pulmonary, diabetes uncomplicated, diabetes complicated, hypothyroidism, renal failure, liver disease, peptic ulcer, aids, lymphoma, metastatic cancer, solid tumor, rheumatoid arthritis, coagulopathy, obesity, weight loss, fluid electrolyte, blood loss anemia, deficiency anemias, alcohol abuse, drug abuse, psychoses, depression. These features are extracted by following the comorbidity concept introduced by the official MIMIC release (\citeauthor{johnson2020mimic}). 

Finally, we convert each patient stay episodes into a mortality prediction dataset. The inputs of the dataset are the combination of time-invariant features of patient demography and comorbidity and time-variant events happened during patient's ICU stay. For the time-variant features, we select 38 signals that have the most occurrences among all the valid ICU stays. Among these 38 signals, heart rate has the most occurrences. We also calculated a relative frequency \footnote{These numbers have no unit, and the data needs to be further inspected, might subject to change later.} of each time-variant feature of the last 24 hours of each ICU stay in Table \ref{tab:variables}. The relative frequency for each signal is calculated based on the occurrences of the signal divided by the Basophils occurrences.
The labels of the dataset are indicators of whether the patient passed away after their ICU stay.  
In order to perform our irregular time-series patient mortality classification, we have to check whether each ICU stay's end time is before the record time of death of the patients. In order for our model to learn meaningful representation, we also eliminated the ICU stays with duration less than 12 hours and stays that has less than 5 lab tests ordered.
After preprocessing with these basic criterion of the ICU stays, we selected around $\sim$50,000 patients' $\sim$70,000 ICU stays. The morality rate of the total stays is around 12\%. 

\begin{table}[h]
\centering
\caption{List of Time-Variant Features and their Relative Frequencies}

\begin{tabular}{@{}ll@{}}
\toprule
\textbf{Variable} & \textbf{Relative Freq} \\
\midrule
Anion gap & 0.21 \\
Bicarbonate & 1.18 \\
Blood urea nitrogen & 0.22 \\
Chloride (blood) & 0.22 \\
Creatinine (blood) & 0.59 \\
Diastolic blood pressure & 1.01 \\
Heart Rate & 7.22 \\
Hematocrit & 1.223 \\
Hemoglobin & 6.42 \\
Mean blood pressure & 1.15 \\
Mean corpuscular hemoglobin & 1.48 \\
Platelets & 1.43 \\
Potassium & 1.12 \\
Red blood cell count (blood) & 0.24 \\
Sodium & 1.22 \\
Systolic blood pressure & 6.35 \\
White blood cell count (blood) & 0.29 \\
Albumin & 0.24 \\
Alkaline phosphate & 1.12 \\
Basophils & 1 \\\
Bilirubin (total) & 1.34 \\
CO2 (ETCO2, PCO2, etc.) & 0.59 \\
Calcium (total) & 0.8 \\
Calcium ionized & 1.33 \\
Eosinophils & 0.37 \\
Lactate & 0.68 \\
Lactic acid & 1.12 \\
Magnesium & 0.29 \\
Monocytes & 0.23 \\
Partial pressure of carbon dioxide & 1.03 \\
Partial thromboplastin time & 1.16 \\
Prothrombin time & 0.23 \\
pH (blood) & 0.09 \\\
Asparate aminotransferase & 0.75\\
Oxygen saturation & 3.87 \\
Phosphate & 6.62 \\
Fraction inspired oxygen & 6.43 \\
Temperature (C) & 1.11 \\
\bottomrule
\end{tabular}
\label{tab:variables}
\end{table}

% We consider time-invariant features as ...

% We consider time-variant features as ...

\section{RL Experience Generation} \label{apd:second}

%RL is often studied for training an agent that observes the current state of the environment and takes actions to maximize its reward to transit to the next state. 
%Deep-RL is suitable for problems that require decision-making, e.g., medical decisions. Researchers have been applying Deep-RL to find policies for treatment plans \citeauthor{ShamimNemati2016OptimalMD, AniruddhRaghu2017ContinuousSM, NiranjaniPrasad2017ARL, JosephFutoma2018LearningTT}. However, only two works have considered RL for lab test scheduling \citeauthor{cheng2018optimal, chang2019dynamic}.

\subsection{Modeling Patient Trajectory (Mortality)} \label{lstm}
For each patient's ICU stay, we assume that we have the patient's time-invariant information like diagnosis, comorbidity, and demographic information represented by a vector $X_{inv} \in \mathbb{R}^u$ where $u$ is the number of features needed to represent patient time-invariant information. 

Since different signals and patient lab test measurements are collected at different frequency across the entire duration of the ICU stay, we first discretize the ICU stays into various time intervals. 
Within each time interval $t$, for each possible time-variant feature (signal), we average over all the values measured during this time interval. 
For instance, if time-interval $t$ represents patient ICU stay between hour 3 and 4, then if a signal $x_i$ has two measurements at time $3.1$ and $3.5$, then $x_i^t = (x_i^{3.1} + x_i^{3.5})/2$. We define our time-variant feature based on time interval $t$, and $X^t_{tv} = [X_1^t, X_2^t,...,X_m^t] \in \mathbb{R}^m$ where $m$ is the number of time-variant signals we consider from our dataset.  

For each ICU stay, we also know the patient mortality $y \in \{0,1\}$\footnote{Since MIMIC dataset stops collect data 6 hours before the patient dies, we don't have to worry about data leakage issue.} after the stay. After processing the raw MIMIC-IV data, we get a dataset $\mathcal{D} = \{(X_{inv}, X_{tv}, y)_1, (X_{inv}, X_{tv}, y)_2, ..., (X_{inv}, X_{tv}, y)_n\}$ where $n$ is the number of ICU stays in the dataset. Here, each $X_{tv}$ is a $T$ by $m$ matrix 
$$X_{tv} = \begin{bmatrix}
X_1^0& X_2^0 &X_3^0& ... &X_m^0\\
X_1^1& X_2^1 &X_3^1& ... &X_m^1\\
...\\
X_1^T& X_2^T &X_3^T& ... &X_m^T\\ 
\end{bmatrix} \in \mathbb{R}^{T \times m},$$
where $T$ is the total number of time-intervals considered for each stay and $m$ is the number of time-variant signals we consider from our dataset. For our project, we choose $T = 23$ and $m = 38$ for all the ICU stays in $\mathcal{D}$.

In order to obtain a representation of patient status for each time interval $t$, we first use the dataset $\mathcal{D}$ to train a time-series model that performs a binary classification task to predict patient's mortality within the next 24 hours. For each ICU stay $(X_{inv}, X_{tv}, y)_i \in \mathcal{D}$, we duplicate the time-invariant vector $X_{inv}$ $T$-times and concatenate this $T$ by $u$ matrix to the $X_{tv}$ to obtain
$$X = \begin{bmatrix}
X_1^0 \text{ } ... &X_m^0 &X_{inv, 1} &  ...  \text{ } X_{inv, u}\\
X_1^1  \text{ }  ... &X_m^1 &X_{inv, 1} & ...\text{ }  X_{inv, u}\\
&\text{ } ...\\
X_1^T \text{ } ... &X_m^T &X_{inv, 1} & ... \text{ } X_{inv, u}\\ 
\end{bmatrix} \in \mathbb{R}^{T \times (m+u)},$$ 
where $X$ is the input for our patient trajectory model $f(X) : \mathbb{R}^{T \times (m+u)} \rightarrow \{0,1\}$. The patient trajectory model $f(X)$ takes a input matrix that represents a patient ICU stay and outputs the patient mortality within the next 24 hours. We choose to use LSTM (\citeauthor{hochreiter1997long} as our $f(X)$) and apply gradient-based supervised learning methods to train our $f(X)$. 

By defining our patient trajectory model this way, we hope that the hidden state of the trained LSTM (\citeauthor{hochreiter1997long}) model captures the patient status up to time $t$ when we construct $X$ such that last row of $X$ is the patient ICU stay data at time $t$. 

\subsection{RL Experience Preparation} 

With the trained patient trajectory model, we are able to prepare the experience tuple $(S, A, R, S')$ for our offline deep-RL policy learning.
Since Chang et al.'s method \citeauthor{chang2019dynamic} is not confined by any specific disease, we can define state and reward in a more general way than the prior work. 
After building a binary classifier with a LSTM model that predicts whether the patient will pass away within the next 24 hours, for each time interval $t$, we can obtain hidden state $h_t$. 
We then construct our state vector as $$s_v = [h_t, \text{multihot}(A^v)],$$ where $h_t$ is the last hidden state of the trained LSTM at time-interval $t$ to represent patient history and $\text{multihot}(A^v)$ is the history of lab tests has been measured up to step $v$ for this patient.

The subscript $v$ is introduced here since a special choice of \citeauthor{chang2019dynamic}'s work on constructing the action space. By our prior formulation, each action at each time step should be a binary vector of length $K$ where $K$ is the total number of tests that can be ordered. Under this setting, the action space has $2^K$ total possible combinations. When $K$ is getting bigger, the computational cost is extremely high. Since we consider $m = 38$ for our project, we have to modify the policy action space. 

We do this by introducing index $v$ which represents different lab test ordered between time $t$ and $t+1$. By this modification, our policy $\pi : s_v \rightarrow a$ only outputs one action each step. To indicate the stop of ordering lab tests between time $t$ and $t+1$, we introduce a new action $\Omega$ as a stop action to form our action space $A = \{1, 2, 3, ..., m, \Omega\}$. For each step $v$, the action is just a single element from all possible actions $A$. Algorithm \ref{alg:getaciton} shows how to run our policy under this updated setting. 

After defining state and action, we define the reward as the combination of information gain and action cost for each step $v$
$$r(s_v, a_v) = \Delta p - \lambda * c(a_v), \text{ where}$$
\begin{equation}
  \Delta p =\begin{cases}
    f_p(X_v)  - f_p(X_{v-1}), & \text{if $y = 1$}\\
    - (f_p(X_v)  - f_p(X_{v-1})), & \text{otherwise}
  \end{cases}, \text { and }
\end{equation}
\begin{equation}
  c(a_v)=\begin{cases}
    1, & \text{if $a_v \neq \Omega$}\\
    0, & \text{otherwise}
  \end{cases}
\end{equation}
Here, $\lambda$ is a hyperparameter that represents the cost coefficient and $c(\cdot)$ is the cost of a action. For now, we treat each action with same cost. $\Delta p$ represents the information gain which is the probability change of our classifier $f$ before and after the lab test $v$ is measured at time $t$. The $X_v$ and $X_{v-1}$ differs by the value that lab test that step $v$ measures. This means that $X_v$ and $X_{v-1}$ differs on an element on the last row of the matrix $X^t$. 

With the definition of action, states and reward at each step $v$ of each $t$, for each ICU stay, we can generate two types of RL experience tuples: 1) transition between each time step $t$ (index by $v$) which we refer to as per-step experience, and 2) transition between time step $t$ and $t+1$ which we refer to as time-passing experience. Our RL experience generation can be described by Algorithm \ref{alg:getrlexp}.
\begin{algorithm2e}
\caption{Generate RL experience for a Patient ICU stay at time $t$}
\label{alg:getrlexp}
\SetKwProg{generate}{Function \emph{generate}}{}{end}
\SetKwRepeat{Do}{do}{while}
%\SetKwProg{generate}{Function \emph{generate}}{}{end}
\KwIn{Pretrained LSTM model $f$, Lab tests ordered between $t$ and $t+1$: $A' \subseteq A$, Lab tests ordered prior to $t$: $A^t \subseteq A$} %and their values $X_{inv}^t = [x_1^t, ..., x_m^t ] $ between $t$ to $t+1$, }
\KwOut{All training experience tuples $E$}
\nosemic $E = \emptyset$ \;
\nosemic Store time-passing experience from $t$ to $t+1$\;
\nosemic  $(s, a, r, s')_{\text{time-pass}} = ([h_t, \mathrm{multihot}(A^t \cup A')], \Omega, f_p(X_{t+1}) - f_p(X_t), [h_{t+1}, \emptyset] )$ in $E$\;
\nosemic Randomly shuffle $A'$\; %list($A'$)\;
\For{$v = 1$ \KwTo $|A'|$}{
Store per-step action experience \;
$(s,a,r,s)_{\text{per-step}} = ([h_t, \mathrm{multihot}(\{a'_0,...,a'_{v-1}\}\cup A^t)], a'_v, \Delta p - \lambda \cdot \mathrm{cost}(a'_v),   [h_t, \mathrm{multihot}(\{a'_0,...,a'_{v}\}\cup A^t)])$ in $E$  
}
\end{algorithm2e}

\section{RL Algorithms} \label{apd:third} 
Given a policy $Q$, the policy is defined by Algorithm \ref{alg:getaciton}.
% \begin{algorithm2e}
% \caption{Computing Net Activation}
% \label{alg:net}
%  % older versions of algorithm2e have \dontprintsemicolon instead
%  % of the following:
%  %\DontPrintSemicolon
%  % older versions of algorithm2e have \linesnumbered instead of the
%  % following:
%  %\LinesNumbered
% \KwIn{$x_1, \ldots, x_n, w_1, \ldots, w_n$}
% \KwOut{$y$, the net activation}
% $y\leftarrow 0$\;
% \For{$i\leftarrow 1$ \KwTo $n$}{
%   $y \leftarrow y + w_i*x_i$\;
% }
% \end{algorithm2e}

For Off-policy policy evaluation, with the trained estimator regression model $\phi$ and the dataset $\mathcal{D}$, for each policy $Q$ we calculate the cumulative information gain $G$ on the test set by summing the estimated information gain for each time step $t$ for each ICU stay $X^i \in \mathcal{D}$. This per-time off-policy evaluation is shown in Algorithm \ref{alg:oppe}
.\begin{algorithm2e}
\caption{Running Policy}
\label{alg:getaciton}
\SetKwProg{generate}{Function \emph{generate}}{}{end}
\SetKwRepeat{Do}{do}{while}
%\SetKwProg{generate}{Function \emph{generate}}{}{end}
\KwIn{LSTM hidden state $h_t$, policy $Q$}
\KwOut{DQN actions $A_t$}
\nosemic Initialize actions $A_t = \emptyset$\;
\While{$\Omega \notin A_t$}{
\nosemic $s_t \gets [h_t, \mathrm{multihot}(A_t)]$\;
\nosemic $a \gets  \operatorname{argmax}_{a' \notin A_t} Q(s_t, a')$\;
\text{Add} $a$ \text{into} $A_t$
}
\end{algorithm2e}

\begin{algorithm2e}
\caption{Per-time off-policy evaluation}
\label{alg:oppe}
\SetKwProg{generate}{Function \emph{generate}}{}{end}
\SetKwRepeat{Do}{do}{while}
%\SetKwProg{generate}{Function \emph{generate}}{}{end}
\KwIn{Trained value estimator regression model $\phi$, patient database $D = \{X^1, X^2, ..., X^N\}$, DQN state $s_t^i$, trained DQN agent $Q$}
%Pretrained LSTM model $f$, Lab tests ordered $A' \subseteq A$ and their values $X_{inv}^t = [x_1^t, ..., x_m^t ] $ between $t$ to $t+1$, }
\KwOut{Estimated cumulative information gain $G$}
\nosemic $G = 0$ \;
\For{$X^i \text{ in } D$ }{
    \For{$t=1$ \KwTo $T^i$ }{
    $a_t^Q \gets$ run $Q$ with patient state $s_t^i$ (Algo \ref{alg:getaciton}) \;
    $\Delta_p^Q = \phi(s_t, a_t^Q)$ \text{// Estimate probability changes} \;
    $G = G + \gamma * (-1)^{y^i+1}\Delta_p^Q$ 
    }
% Store per-step action experience \;
%$(s,a,r,s)_{\text{per-step}} = ([h_t, X^t_{tv, v-1}], \text{list}(A')[v], \Delta p - \lambda \cdot 1,   [h_t, X^t_{tv, v}])$ in $E$  
}
\end{algorithm2e}

\section{Other Results} \label{apd:fourth}
For each method we experimented, we show the policies with highest information gain with lowest cost in Figure \ref{fig:m3} and \ref{fig:m4}. 
For MIMIC-III, the physician policy cost $c$ and information gain $g$, $(c, g) = (10.16, -0.023)$. The values of each methods are shown in Table \ref{tab:m3vals} .
For MIMIC-IV, the physician policy cost $c$ and information gain $g$, $(c, g) = (9.27, -0.023)$. The values of each methods are shown in Table \ref{tab:m4vals}.

% \begin{table*}[htbp]
% \floatconts
%   {tab:operatornames}%
%   {\caption{Predefined Operator Names (taken from 
%    \textsf{amsmath} documentation)}}%
%   {%
% \begin{tabular}{rlrlrlrl}
% \cs{arccos} & $\arccos$ &  \cs{deg} & $\deg$ &  \cs{lg} & $\lg$ &  \cs{projlim} & $\projlim$ \\
% \cs{arcsin} & $\arcsin$ &  \cs{det} & $\det$ &  \cs{lim} & $\lim$ &  \cs{sec} & $\sec$ \\
% \cs{arctan} & $\arctan$ &  \cs{dim} & $\dim$ &  \cs{liminf} & $\liminf$ &  \cs{sin} & $\sin$ \\
% \cs{arg} & $\arg$ &  \cs{exp} & $\exp$ &  \cs{limsup} & $\limsup$ &  \cs{sinh} & $\sinh$ \\
% \cs{cos} & $\cos$ &  \cs{gcd} & $\gcd$ &  \cs{ln} & $\ln$ &  \cs{sup} & $\sup$ \\
% \cs{cosh} & $\cosh$ &  \cs{hom} & $\hom$ &  \cs{log} & $\log$ &  \cs{tan} & $\tan$ \\
% \cs{cot} & $\cot$ &  \cs{inf} & $\inf$ &  \cs{max} & $\max$ &  \cs{tanh} & $\tanh$ \\
% \cs{coth} & $\coth$ &  \cs{injlim} & $\injlim$ &  \cs{min} & $\min$ \\
% \cs{csc} & $\csc$ &  \cs{ker} & $\ker$ &  \cs{Pr} & $\Pr$
% \end{tabular}\par
% \begin{tabular}{rlrl}
% \cs{varlimsup} & $\varlimsup$ 
% & \cs{varinjlim} & $\varinjlim$\\
% \cs{varliminf} & $\varliminf$ 
% & \cs{varprojlim} & $\varprojlim$
% \end{tabular}
% }
% \end{table*}

\begin{table*}[htbp]
    %\centering
    %\captionsetup{justification=centering,margin=2cm}
    \caption{Exact Values of Cost and Information  for Different Methods in Figure \ref{fig:m3}}
    \begin{tabular}{c|c}
    \hlineB{2}  % Thicker line
        \textbf{Method} & (Policy Cost - C, Cumulative Gain - G) \\
        \hlineB{2}  % Thicker line
        Physician & $(10.16, -0.023)$\\
        BC &  $(2.49, -0.024)$, $(9.57, -0.019)$, $(64.99, -0.014)$, $(149.86, 0.015)$, $(174.16, 0.026)$, $(271.22, 0.05)$\\
        DDQN & $(0.34, -0.018)$, $(84.67, -0.01)$, $(172.57, 0.082)$   \\
        CQL & $(65.81, 0.002)$, $(65.98, 0.046)$, $(351.93, 0.067)$   \\
        IQL & $(0.01, -0.018)$, $(44.63, -0.008)$, $(111.63, 0.048)$, $(400.39, 0.105)$ \\
    \hlineB{2}
    \end{tabular}
    \label{tab:m3vals}
\end{table*}

\begin{table*}[htbp]
    %\centering
    \caption{Exact Values of Cost and Information  for Different Methods in Figure \ref{fig:m4}}
    \begin{tabular}{c|c}
    \hlineB{2}  % Thicker line
        \textbf{Method} & (Policy Cost - C, Cumulative Gain - G) \\
        \hlineB{2}  % Thicker line
        Physician & $(9.27, -0.023)$\\
        BC &  $(2.49, -0.024)$, $(9.57, -0.019)$, $(64.99, -0.014)$, $(149.86, 0.015)$, $(174.16, 0.026)$, $(271.22, 0.05)$\\
        DDQN & $(0.34, -0.018)$, $(84.67, -0.01)$, $(172.57, 0.082)$   \\
        CQL & $(65.81, 0.002)$, $(65.98, 0.046)$, $(351.93, 0.067)$   \\
        IQL & $(0.01, -0.018)$, $(44.63, -0.008)$, $(111.63, 0.048)$, $(400.39, 0.105)$ \\
    \hlineB{2}
    \end{tabular}
    \label{tab:m4vals}
\end{table*}

We only report the best values for each method in Figure \ref{fig:m3} \& \ref{fig:m4}. However, we trained various policies for each method. Based on the intervals of relative cost, we show the distribution of each method's information gain in Figure \ref{fig:apdm3} \& \ref{fig:apdm4}. We find that behavior cloning performs better than DDQN when the cost is low for both datasets. For MIMIC-IV, CQL performs the best among all methods when the relative cost is low. The reproduced result for DDQN follows the result obtained from \citeauthor{chang2019dynamic}.

\begin{figure}[h]
\floatconts
  {fig:apdm3}
  {
  \vspace*{-1.5\baselineskip}
  \caption{Box-whistle plots of all policies in MIMIC3}}
  {%
    \includegraphics[width=1\linewidth]{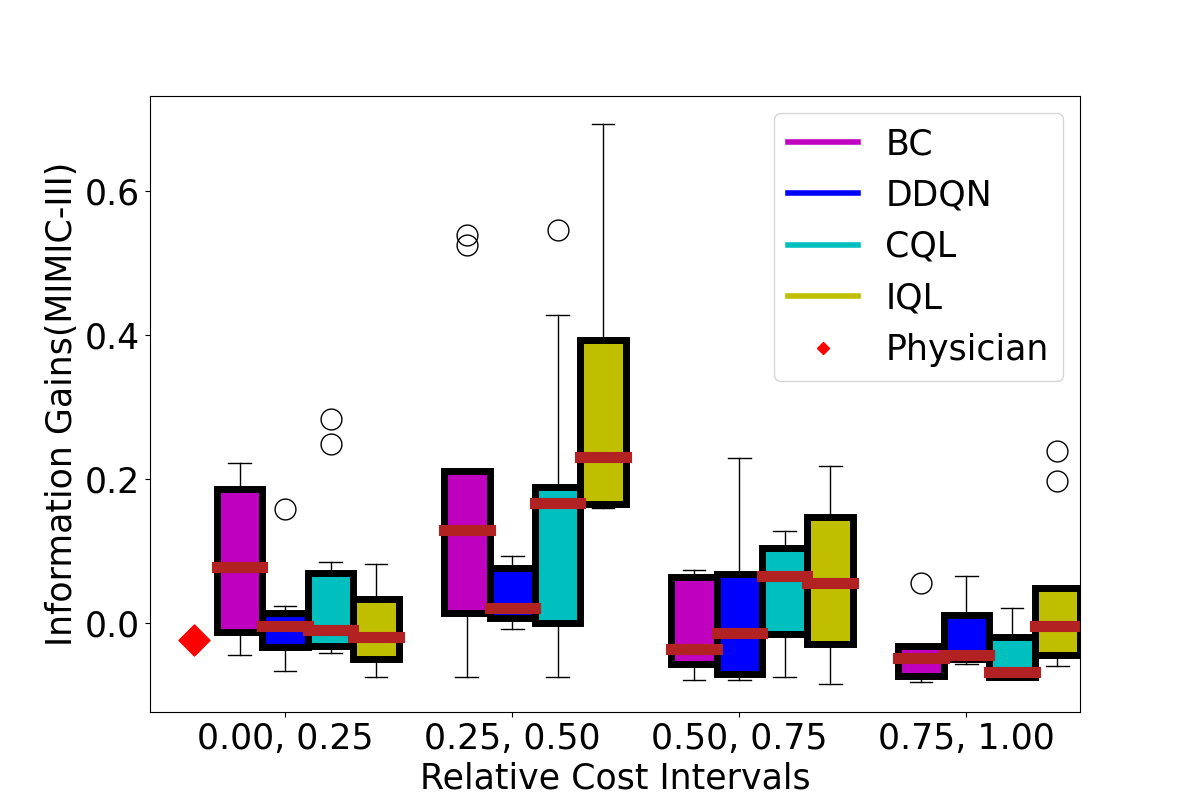}
  }
\end{figure}

\begin{figure}[h]
\floatconts
  {fig:apdm4}
  {
  \vspace*{-1.5\baselineskip}
  \caption{Box-whistle plots of all policies in MIMIC4}}
  {%
    \includegraphics[width=1\linewidth]{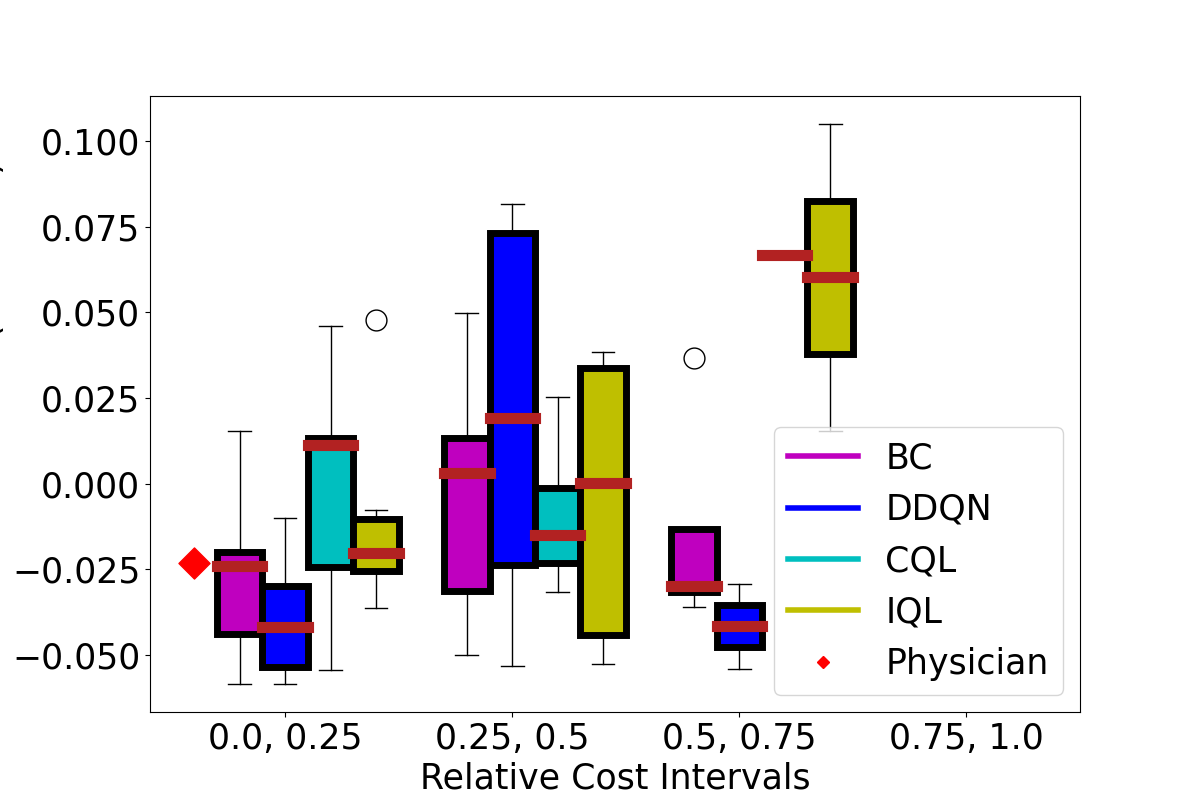}
  }
\end{figure}

\end{document}